## A Study of Sindhi Related and Arabic Script Adapted languages Recognition

D. N. HAKRO[++], A. Z. TALIB*, Z. BHATTI**, G. N. MOJAI**

School of Computer Science, University Sains Malaysia, PO box 11800, USM Penang, Malaysia

**Abstract:** A large number of publications are available for the Optical Character Recognition (OCR). Significant researches, as well as articles are present for the Latin, Chinese and Japanese scripts. Arabic script is also one of mature script from OCR perspective. The adaptive languages which share Arabic script or its extended characters; still lacking the OCRs for their language. In this paper we present the efforts of researchers on Arabic and its related and adapted languages. This survey is organized in different sections, in which introduction is followed by properties of Sindhi Language. OCR process techniques and methods used by various researchers are presented. The last section is dedicated for future work and conclusion is also discussed.

## 1. INTRODUCTION

The character recognition of the Roman type of languages especially English has come near to perfection and it is also considered as one of the successful application in the field of computer vision. The work on Arabic script and other scripts is being continued on; but the languages adopting Arabic script is very little while the work on Sindhi language is near to its origin. The Arabic script has more complexities as it's an entirely different script as compared to Roman script. A significant work also has been done on Indian local Languages (Indian Scripts) (Pal and Chaudhuri, 2004; Hewavitharana and Fernando, 2002; Chaudhuri *et al.,* 2002; Sharma and Jhajj, 2010; Lakshmi *et al.,* 2006; Bansal and Sinha, 2002; Basu *et al.,* 2009) but Sindhi is lacking its fully functional OCR, although the remarkable work has been done on Sindhi Computing(Bhatti *et al.,* 2014). This paper presents a review of the character recognition processes and image processing techniques applied in OCR systems. The techniques include text line segmentation, word and character segmentation and classification. The paper also looks in to the choices of researchers, made for their research in various languages all around the world.

## 2. PROPERTIES OF SINDHI LANGUAGE

According to Moulana Ubedullah Sindhi a well know Islamic Philosopher and Scholar wrote in his book about Sindhi language, "The seven languages are the main languages in which Holy Books were sent and the remaining world languages are derived from these seven languages. Sindhi is one these languages with Arabic and Hebrew" (Allana, 2004). The rich historical background of Sindhi language can be inferred from the 5000 years Indus Civilization of Moen-jo-Daro near Larkana district of Sindh (AboutIndus, 2014). In (Al-lana, 2004) Dr. Nabi Bux Khan Baloch; a well-known Sindhi historian and scholar has categorized Originity of Sindhi into different opinions in which one of the opinion explains that Sindhi is a Sanskrit branch via Varchada Apabharansha.

Sindhi Language is spoken by 18 million people in Pakistan as well as 2.8 million in India. Two common scripts, Arabic and Devanagri are used for writing Sindhi language. Arabic is the most common script used, by adding some modified letters to the Arabic letters. In India Sindhi is written in both scripts because it can also be written with Devanagri script. Sindhi Language has 24 more letters (total 52) than Arabic language with 28; some modified letters have been added with four dots to accommodate the different sounds. Sindhi has more vowels and consonant than Arabic and its neighbor language Urdu. The writing system follows the same style of Arabic script in which letters are written from right to left while the numbers are written from left to right.

A sum total of 52 characters, in which some characters may have their two to four shapes according to their position in a sentence shown in **(Fig.1)**. Characters are connected to make a component, like Arabic script and its adopting languages such as Persian, Urdu and Pashto.

| Isolated | Start | Middle | End |
|----------|-------|--------|-----|
| ٻ | ٻـ | ـٻـ | ـٻ |
| ڠ | ڠـ | ـڠـ | ـڠ |
| ڱ | ڱـ | ـڱـ | ـڱ |

Fig. 1: Different forms of letters according to their positions

[++]Corresponding author E-mail: dn11_com135@student.usm.my
*Institute of Information and Communication Technology, University of Sindh, Jamshoro, Pakistan
**Institute of Mathematics and Computer Science (Bioinformatics), University of Sindh, Jamshoro, Pakistan

According to shapes different classes can be made and these classes have same base and differ in number, position and placement of dots as shown in **(Fig.2)**. These main difficulties, Issues and Challenges related to Sindhi OCR especially in segmentation and recognition of Sindhi characters are identified and discussed in Hakro *et al.,* 2014.

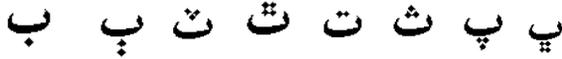

**Fig. 2: Same base with different number of dots and their positions**

3.                    **HISTORY OF OCR**

After the invention of computer and data processing by computers, it was necessary to process the data with a great speed and less efforts. Data processing with computers was performed through the punched card, which was a time consuming and error prone. This was the birth of OCR and the first commercial OCR was available in mid 1950s (About OCR, 2014). With the advancement of machine reading and Artificial Intelligence OCR became the point of interest for the researchers around the world. After successful maturity of Roman OCRs, the new ways were open for the other scripts. This led also towards the development of OCRs for the specific language and then researchers were directed to multifont, omnifont, Multilanguage and high performance OCRs.

4.                    **OCR TECHNIQUES**

In this section we present the review of the techniques applied by various researchers around the world for various languages and scripts. OCR converts a text image to a text file which contains the characters which are available for editing. These characters are represented in ASCII or UNIcode while the previous text image characters were represented in image by a position of pixel elements. In a simple way the OCR converts a text image to a text file.

5.      **OCR Elements**

The general diagram of the OCR is showing below in **(Fig. 3)**.

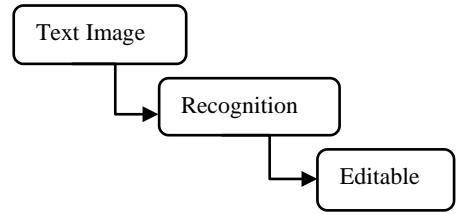

**Fig. 3**: **General Process of OCR**

The image is scanned and converted (Digitized) into bitmap. The converted image is passed through the preprocessing techniques. The preprocessing techniques are applied according to different conditions such as scanning errors, text with other objects, text skewed in scanning or lines of text, noise available, font, size and script of the text. These preprocessing techniques are selected by researchers on demand of their research and limitations as well as the need of time. The researchers can select one or all of the preprocessing techniques. The common preprocessing techniques step in OCR is shown in **(Fig.4)**. After preprocessing the characters are segmented from lines, words as well as characters which will be recognized with the help of matching with the feature vector. In classification, on the success of matching the characters, corresponding code is sent to a text file, which will be the ultimate output of the scanned image. General steps of OCR are given below in **(Fig.5)**
.

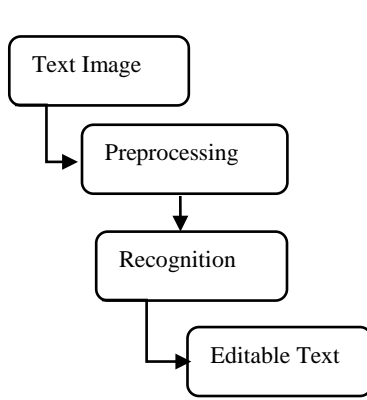

**Fig. 4: Preprocessing**

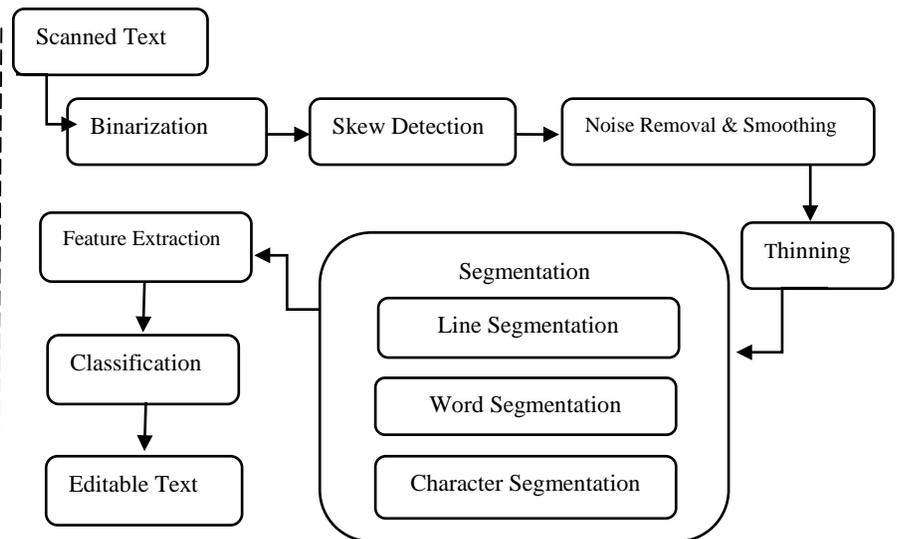

**Fig. 5: General steps in OCR**

## 5.1    Binarization

After scanning of the text image, it is converted from color or grayscale and then binary format which is of two colors, black and white. The two colors represent the foreground and background and become easy to extract the text from the image. The difference between the gray scale image and binary is the range of values in gray scale; there are 256 levels of white and black whereas the binary contains only two values. For binarization any value can be selected (threshold value) and the above level values are given to white scale while below values are given to black scale which forms binary image. (Pechwitz *et al.,* 2002) presented Arabic database containing the names of villages handwritten by 411 subjects. The INF/ENIT data base contains 26400 naming words consists of 210000 characters manually segmented and converted into binary form. and reported 89% of maximum word level accuracy. Ajward *et al.* (2010) employed the binarization for the recognition of Sinhala text.

(Ashwin and Sastry, 2002) used global threshold to convert an image scanned at 300 DPI for their size and font independent recognition of Khannada characters, Basu *et al.* 2009 employed threshold to convert their feature set for their handwritten Bangla characters. **(Fig. 6)** illustrates the binarization process in which Sindhi alphabet characters are converted (binarized) in to two tone binary image. (Pechwitz & Maergner, 2003) Performed various tests on INF/ENIT data base which is freely available

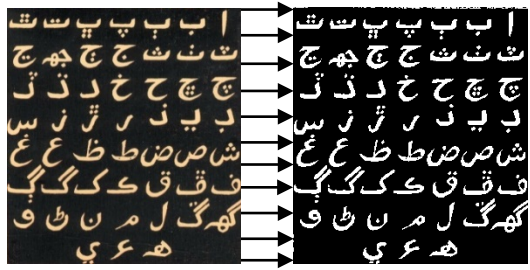

(a) Gray Scale Image  (b) Binary image (binarized)
**Fig.6: Binarization of Sindhi Characters**

## 5.2    Skew detection

The OCR processes are working towards perfection if the scanned document is given in a right direction so for this is necessary to scan document with care. The errors in scanning or error produced by any other reason can produce some diverted or slanted page. To detect this slant or skew of the text lines, which makes text in image skewed, the angle of skewness is identified and it is corrected. (An-Shatnawi and Omar, 2009) proposed an arbitrary polygon method based derivation which uses centroid baseline of polygon. The authors reported 87% of accuracy on Arabic documents. The

Arabic documents were scanned from 150 various newspapers, books, handwritten documents and journals. The process is divided into two parts the first one is baseline identification in which center of gravity or center of mass has been taken as the centroid. The angle is measured with respect to the rotation angle by which text is in readable form. Chaudhuri *et al.* (1997) proposed a skew detection and correction method for Indian script documents whereas Desai, (2010) detected skew and corrected on individual numbers of Guajarati script. In their work, $10^0$ angles are used for experiments and created 5 more shapes of individual characters each with the difference of $2^0$ angles. **(Fig. 7)** shows the skewed image and skew corrected document.

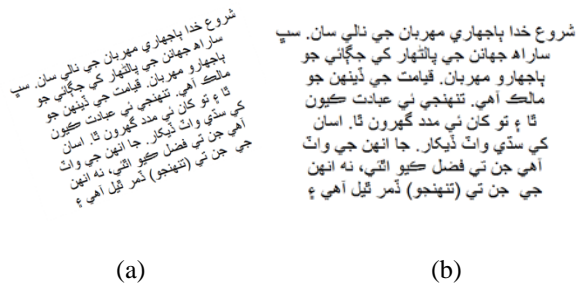

(a)                              (b)

**Fig. 7: (a) skewed image (b) corrected image**

## 5.3    Noise Removal and smoothing

The popular name of the noise is unwanted signals. The noise is anything that is not present in the object of the image and some undesired or unwanted lines, dots on the scanned page. This noise can occur because of the quality of the paper, printing ink, printer or any other source and removed by employing some special noise removal techniques and algorithms. The noise on an image can pose difficulties in recognition as well other processes and this becomes worse when the OCR process is applied in Arabic script or its adopting languages due to the resemblance of noise dots with the character dots. It is very difficult to differentiate between the noise and the constitution of the object in an image (Introduction to Image noise, 2013). The pixels can be estimated by only their position that it is a noise or a part of an object. During OCR preprocessing some researchers follow the image noise removal and some assume the noise free images (Ajward *et al.* 2010) on a high quality paper for their experiments.

## 5.4    Normalization

Normalization is also used in preprocessing which reduces the size of character as well ensures the character is in center (Haj-Hassan,1991; Mahdi,1989; Abbas *et al.* (1986)). When OCR system is oriented and sensitive to size and font, then normalization is helpful to overcome problem of size variations especially in correlation and template matching (Al-Badr and Mahmoud, 1995). (Fakir and Sodeyama, 1993) employed normalization to their segmented characters to fit in a certain size of the box. The expansion of the characters is performed by x and y factors independently. The horizontal projection is employed by the Haj-Hassan (1985), followed by Jambi (1992) for the estimation of the width average to segment the character as well as for font recognition.

### 5.5 Thinning

Cowell and Fiaz, (1992) identified that a simple search on the internet generates 150 papers on thinning algorithm but this number is even higher than once reported. The importance of thinning can be understood very well from this number. Thinning is a basic operation used for making a shape, object or any structure reduce at the size of one pixel thick. This operation is used in Optical Character Recognition as well as some other applications (Gonzalez *et al.,* 2005). The feature extraction information is distinctive and processing of these features takes too much time and efforts for the machines. To reduce this significant amount of information for processing thinning is used to reduce the amount of information and process with ease. Thinning algorithms, when applied can produce the noise to some extent when applied to software or hardware (Cowell and Fiaz ,1992). Skeletonisation or thinning is used in various literatures. Lam *et al.* (1992) Reports that circuit boards were designed on the basis of recognition on thinned characters performed by Sherman, Duetsch and Alcorn in 1959, 1968 and 1969 respectively. Various applications such as blood cells analysis, quantitative metallography, circuit boards, have been reported in Lam *et al.* (1992). Optical character recognition has become main application where thinning algorithm is used to thin the skeleton of character so that a character can be processed easier. Shang and Yi, (2007) proposed inner and outer image based region filling technique to use with binary images applying pulse coupled neural network. Shang claims that his algorithm is potential for exploring new applications. Shang used firing stage in which final produced thinned result is decided and on failure the process is repeated further till final thinned image. Jagna and Kamakshiprasad, (2010) proposed a 3x3 mask pattern for the pixel deletion in a parallel algorithm which also preserves the connectivity. In (Zhang and Suen, 1984), two sub-iterations based fast algorithm is proposed which deletes the boundary points. These boundary points are north-west points and south-east boundary points. In (Chiu and Tseng,1997), feature preservation based thinning algorithm containing of various stages is proposed for the handwritten Chinese characters. Direction codes are used to decide the direction of pixels or connectivity. The other segment is divided in fork segments as well as in strokes. The third phase is dedicated for stroke extraction and the removing of hair branches and noise. The final phase connects the extracted strokes in previous step. (Aghbari and Brook, 2009) claimed that the classical thinning algorithm of Hilditch, (1983) does not perform well due to the nature of Arabic script and they have developed their own thinning algorithm for their Arabic OCR based on the same idea of Hilditch, (1983) and claim that their algorithm retains the Arabic shape and loops of the Arabic characters. Figure 8 illustrates the thinning of Sindhi words.

### 5.6 Segmentation

The image document is scanned, noise removed, skew corrected will now be ready for extracting the readable information, for this extraction, segmentation is the first step. The segmentation includes line, word and character segmentation. The roman script is easy to segment, because characters are not connected and there is no concept of connected words in Roman script such as in English OCR. In Arabic and its script adapting languages, line segmentation, word segmentation and character segmentation is necessary because the words are changing their shape according to their position and connected to each other to form a word. For Arabic script segmentation horizontal projection is the common method. (Pal and Sarkar, 2003) used common projection method to segment the text lines of Urdu text an Arabic script adopting language shown in **(Fig. 8).**

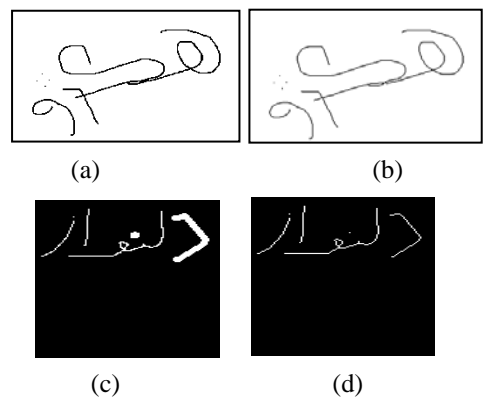

(a)                              (b)

(c)                              (d)

**Fig. 8: Thinning: (a) & (c) original image (b) & (d) Thinned image**

The text lines or the block of lines are segmented by finding out the number of black pixels in a line which build up the valleys of projection profile. The free space between two text lines is the indicators of the separation of the lines of text in an image. A boundary between the two consecutive lines is the indicator of a text line is present (Pal and Sarkar, 2003). Character segmentation is performed by vertical projection profile and the other combined technique component labeling. The two or less than two pixels are represented as the value 0 so that in vertical projection and the else is used to count the number of pixels in an object (Pal and Sarkar, 2003). (Aghbari and Brook, 2009) used projection technique for the segmentation of text lines, words and characters in three steps. The first one segments the text lines from the Arabic manuscript image. In the second step the line image is segmented in to word images by locating free spaces between the words (inter-word space). The same vertical projection is used to segment the connected parts of the word in third and last step of segmentation. (Al-Badr and Mahmoud, 1995) claims in his survey that the cursive Latin segmentation methods can work with Arabic but generally these methods are not suitable for the Arabic script. Omidyeganeh *et al.* (2005) proposed a new segmentation algorithm for 6 various fonts in which documents scanned at 300 dpi and stored in two tone images. The text lines of these images are segmented by projection profile. Their segmentation algorithm is based on conditional labeling in which up contours and down contours are labeled. These contour curvatures are grouped in the next step of their algorithm. These up and downward contours are traced and represented with eight directional freeman codes from 0 to 7. The next step is character segmentation based on contour grouping and contour labeled with down, median and up. By dividing the length of subwords to the segmentation points, the adaptive local base line is detected. The algorithm also needs the post processing because some of the characters are oversegmented. (Misbah and Hussain, 2010) segmented Urdu manually cleaned corpora by identifying the boundaries of sequence of ligatures based on statistical model. Word sequence is ranked based on lexical lookup. Top sequence is further processed based on valid words and the final decision is made on probability processing.

Omidyeganeh *et al.* (2005) used conditional labeling by applying down and up contours for their Arabic/Farsi segmentations work on multifont. The contours are labeled with 1, 0 and -1 to the up , middle and down contours according to the distance from the base line. Their results shown 97% accuracy on 22236 characters on testing and no training was adopted. For the intelligent mobile systems Li *et al.* (2011) claim a robust segmentation algorithm for handwritten Uighur language. The conversion of word into strokes starts the process and the strokes are found by using black connected pixels followed by width, height and location. These strokes are analyzed and decided the connection and affiliation of strokes with words. Local high point detection and Harris corner detection are used to identify the potential segmentation points followed by removing of the unwanted strokes. Li *at al.,* (2010) worked on mobile platform for the segmentation algorithm of Uighur language based on hybrid features. Four features such as location based, connectivity based, ownership features and local peak features. Segmentation is done with obtaining subfield by analyzing ownership features and finding out relative position by using peak features and finally the independent segmentation model is obtained. Extensive experiments results are producing 85% accuracy for correct segmentation. Alaei *et al.* (2010) made use of baseline for the segmentation of handwritten Persian characters and claimed that the same algorithm may be used with Arabic as Persian is the Arabic script adopting language. The baseline is traced and additional strokes are removed and the line is detected by conventional projection technique. Expected segmentation points are identified based on local minima and text line mapping. Experiments on 26 subjects written 43 manually segmented text lines and 899 words produced 90.26% accuracy. Erlandson *et al.* (1996) preferred a word level segmentation rather than the characters are segmented for Arabic language. A database containing 48,200 word shapes were created by using feature extracted vectors. The segmented words are compared against these shape vectors in dictionary of known words. (Tolba and Shaddad, 1990) segmented Arabic text by a dedicated window which calculates the segmentation parameters while sliding from right to left. A predefined thresholding value is used to decide after comparing the output parameters of sliding window a region is silent or starting as well as end of the character. The increase in the segmentation parameter is the indicator of starting of a character. Razak *et al.* (2007) presented a segmentation algorithm which works on overlapped Jawi characters in real time implemented in a chip. A region of interest, converted into binary, false local minima is removed and continues until the segment points achieved. The segmentation accuracy is reported 96% and the technique is also compared with other techniques. Mehmet *et al.* (1997) avoided segmentation, so called segmentation free in which characters are extracted and recognized. Features are extracted first and used for training and the probability matrix is calculated from inverse optional vector. The matrix vector is helping in searching and classifying of the characters. The system is also responsible for finding the gaps and removing them. (Javed and Hussain, 2009) segmented Nastalique words by employing overlapping threshold. The conventional method of projection is not suitable that's why they fine-tuned their algorithm according to situation. The reported results show 100% accuracy and on 3655 ligatures with marks associated of which 3436 as successful with accuracy of 94%.

### 5.7    Feature Extraction

Feature extraction is considered the most important part of the any OCR system. An efficient feature extraction method can save a much amount of time as it is considered the base of accuracy. The features are collected in a feature vector and given as input to the clas-sification stage to recognize the characters. The feature extraction stage contains differentiating attributes of the character or the characteristics by which characters can be differentiated. The main human task in feature extraction stage is to invent or select the features which are most suitable for the task of efficient recognition Cheriet *et al.* (2007).

(Zahedi, and Eslami,, 2011) proposed a new type of feature extraction method for recognition of Arabic and Farsi cursive text based on scale invariant feature transform (SIFT features). The SIFT method is robust in dealing with the scale, size and rotation. In first step of SIFT; extreme boundaries and scales are set for the same object to be viewed in different locations. In the next step two images are compared and difference is calculated by difference of Gaussian (DoG). The DoG is applied on the group of re-sampled images to find out the extremes. The feature matching stage is storing the SIFT keys in database. The best match is found using the nearest neighbor and Euclidean distance between the test and training keys. The model verification step is responsible for identifying the group of features using least square method. The last step is used to employ the least square solution in order to discard the outliers.

Mori and Sawaki, (2008) proposed a method for extracting features from web documents as well as video scenes. The method used 67 fonts from 3190 categories of Japanese language for the data set. The higher results are claimed than any other conventional method on video text. Adaptive normalization is the first step of image deformation while feature compensation is the other. The geometric stroke extraction is based on run-length strokes which result in directional strokes. As the input patterns may be corrupted and can cause degradation so to estimate the degradation, feature compensation is used. The process of degradation estimation is done with input pattern run length profile.

(Mori and Sawaki, 2008) proposed a method for handwritten degraded characters based on run-length compensation. Their technique is used to find out directional run-length strokes. The run-length compensation technique is a five stage process which starts with the division of image into NxN local areas. The following step calculates the black pixel run length compensation. The third and middle stage averages these values; the fourth stage calculates the directional contributivety. The final step is multiplying this directional contributivety in their local areas. The method is tested on Kanji charter database ETL-9 (Saito *et al.*1985).

(Al-Taani and Al-Haj, 2010) claimed their online handwritten technique based on structural feature, is efficient or Arabic characters as well as their technique is also adaptable by tablet applications. The system is based on strokes of characters written by mouse considered as one segment and represented by x and y coordinates. Character is drawn in 5x5 for extracting features for classification in which number of segments are the basic parameters. Feature extraction is done with left-right, bottom up, vertical-horizontal ration, density ratio, sharp edges, secondary segment similarity and secondary segments. The experiments performed on ten subjects with five variations with 1400 characters and accuracy reported is 75.3%.Jelodar *et al.* (2005) made a choice of Hit/Miss morphological operator for the feature extraction of the Persian characters. The accuracy is 99.9% due to the testing on only two different size and single font lotus.

## 5.8    Classification

Feature extraction techniques map the features in a feature space while classification labels the points of feature space of class with defined classes (Cheriet *et al.*, 2007). The classification techniques classify the scores (labels) to the defined classes. In other words the classification technique is the last step where feature vectors are compared with the extracted characters and matched and a successful output is produced in a form of editable format. Various classification techniques are used corresponding to the situations such as structural, statistical, Artificial Neural Networks, support vector machines, Morkov models and a combination of different classifiers. Statistical approaches make use of Bayesian theory which reduces classification loss when estimated probabilities and loss matrix are given. The statistical approach can be in parametric or non parametric approaches. Artificial neural networks were designed to mimic the human brain activities and create more active and intelligent machines. Neural networks are used in various fields and proved also suitable for signal processing and pattern recognition. The network model contains neurons as elements to mimic the physical neurons in human body and the interconnections are differentiated according the model of network. These models are recurrent, self-organizing and so on. The network may be single layer or multilayer containing different numbers of nodes (neurons) corresponding situations. Another classifier Support Vector Machines (SVM), a hyper plane classifier which uses statistical theory is also a good choice for pattern recognition.

Generally used as a binary classifier with two classes. Rani *et al.* (2011) made use of SVM for the recognition of 1900 and 2605 English numerals and Punjabi words with varying kernel functions and reported 99.86% accuracy. For the online recognition, structural methods are common as compared to offline character recognition as they make use of fixed dimensions and present as graph, tree or string.

Structural classifier works like human being this is why they record the stroke orientation, combination or character patterns. The two difficulties in structural classifiers are difficulty in extraction of structural components and the learning from the samples. This is reason Artificial Neural Network and statistical methods are used more often than structural classifiers (Cheriet *et al.*, 2007).

Nizamani & Janjua, (2011) Made use of back propagation for the Sindhi characters to be recognized in a drawing panel, where a user is writing with the help of mouse and the drawn character is recognized. A 23 sampled Hidden Markov Model for each Urdu character and fuzzy rules are used by Razzak *et al.* (2009) and reported 87.6% for Nastaliq and 74.1% for Naskh. Nawaz *et al.* (2009) used an xml file for the recognition of Naskh Urdu characters of different fonts and reported 89% accuracy. (Almohri and Gray, 2008) implemented Arabic OCR system in a DSP system classified by neural networks. A fuzzy neural network is used with 14 inputs, 60 hidden layer neurons and one output and produced 95% of accuracy. The accuracy can be increased up to 98% on the use of same font and size of the database. (Amin & Mansoor, 1997) used back propagation trained feed forward neural network for testing of few hundred printed Arabic characters and 98% accuracy is reported at the average speed of 2.87 characters per word. Moradi *et al.* (2010) used MATLAB for their network training, used weights and implemented their system on FPGA. After the implementation on FPGA the accuracy is 96% and on computer it is reported as 97%.

Nawaz *et al.* (2003) proposed an offline Arabic OCR system for single font with variant size and developed in window environment in java programming. The system tested on 200 characters and accuracy reported is 76%. Zheng (2006) employed four classifiers and recognition rate is claimed as 97%. Fakir *et al.* (2000) recognized Arabic characters in two stages in which first main body is classified by dynamic programming and Hough Transform and the other is topological features for complete characters. The recognition is reported 95% accuracy. (Pervez & Mahmoud, 2013) used nearest neighbor for the classification and reported 79.58% of accuracy. Al-Muhtaseb *et al.* (2008) employed Hidden Markov Models on eight different Arabic fonts and the test images are taken from Sahe Muslem and Sahi al-Bukhari books. For their testing of Hidden Makov Models 2500 lines and 266 lines for testing is used. The results are reported for each font separately and 97.86% is given for Andalus and highest accuracy 99.90% in case of Arial font. Pourasad *et al.* (2011) measured Euclidean distance as well as used gradient value of the points for the recognition of Persian OCR, the recognition was performed on character by character. Accuracy is given in range which starts from 91% to 100%. Rani *et al.* (2011) claimed a high accuracy average 99.86% by using SVM employing polynomial kernel functions for the recognition of bilingual OCR which recognizes English numerals and Punjabi words. The data set was selected from 2605 words of Punjabi language and 1900 English numerals.

## 6. ARABIC SCRIPT AND ITS ADAPTING LANGUAGE SYSTEMS

As Sindhi language OCR is going through its initial steps. And lots of efforts are needed to develop a complete OCR system. While a great efforts are involved in Arabic OCR and its adapting languages. The efforts of researchers around the world for the development of Arabic script OCRs and its adapting language OCRs are continued on.

## 7. EXPECTED RESULTS

OCR is a mature field and a lot of work is available on the various language OCRs around the world. Latin script OCRs can be understood as well solved because of their low complexity while the Arabic script and its adapting language still need a lot of consideration to problems due to various level of complexities like different number and position of dots, cursiveness and changing shape of characters corresponding to situation in a sentence. Various researches on Arabic and its adapting languages been highlighted in this paper. Common methods based review of Arabic and its adapting languages were presented. Researchers reported 90% accuracy for the isolated characters whereas the connected character recognition reported ranging from 75% to 100%.

## 8. CONCLUSION AND FUTURE WORK

By using evolutionary techniques, Sindhi language and Arabic script OCR problems can be solved and directed towards its maturity level as the Latin OCRs have already achieved. A high efficiency OCR will be the output of improving conventional techniques of segmentation and feature extraction. Accordingly this paper produced some information about the Arabic and its Adapting languages OCR. Various Arabic script adapting languages, where attention is needed as there are no OCRS for these languages.